\documentclass{article} 
\usepackage{iclr2024_conference,times}

\usepackage{amsmath,amsfonts,bm}
\usepackage{graphicx}
\usepackage{sidecap}
\usepackage{mathrsfs}
\usepackage{multirow, multicol}
\usepackage{caption}

\usepackage{booktabs} 
\usepackage{enumitem}

\usepackage{algorithm}
\usepackage{algorithmic}

\def\eqref#1{equation~\ref{#1}}

\def\1{\bm{1}}

\def\vc{{\bm{c}}}

\def\ve{{\bm{e}}}

\def\vh{{\bm{h}}}

\def\vs{{\bm{s}}}

\def\vx{{\bm{x}}}
\def\vy{{\bm{y}}}

\def\mX{{\bm{X}}}
\def\mY{{\bm{Y}}}

\DeclareMathAlphabet{\mathsfit}{\encodingdefault}{\sfdefault}{m}{sl}
\SetMathAlphabet{\mathsfit}{bold}{\encodingdefault}{\sfdefault}{bx}{n}

\def\gI{{\mathcal{I}}}

\def\gL{{\mathcal{L}}}

\def\gS{{\mathcal{S}}}

\def\gV{{\mathcal{V}}}

\def\gY{{\mathcal{Y}}}

\usepackage{mdframed}
\usepackage{hyperref}
\usepackage{url}

\author{Pengyu Cheng\thanks{Equal Contribution.}  \\
Tencent AI Lab \\
\texttt{pengyucheng@tencent.com}
\And
Ruineng Li$^*$ \\
Columbia University \\
\texttt{rl3315@columbia.edu}
}

\title{Replacing Language Model for Style Transfer}

\usepackage{bibentry}

\usepackage{microtype}
\usepackage{graphicx}
\usepackage{subfigure}
\usepackage{booktabs} 

\iclrfinalcopy

\usepackage{amsmath}
\usepackage{amssymb}
\usepackage{mathtools}
\usepackage{amsthm}

\usepackage[capitalize,noabbrev]{cleveref}

\theoremstyle{plain}

\theoremstyle{definition}

\theoremstyle{remark}

\usepackage[textsize=tiny]{todonotes}

\newcommand{\MASK}{\texttt{[MASK]} }

\newcommand{\PAD}{\texttt{[PAD]} }

\begin{document}

\maketitle

\begin{abstract}
\vspace{-2mm}
We introduce \textit{replacing language model} (RLM), a sequence-to-sequence language modeling framework for text style transfer (TST). Our method \textit{autoregressively} replaces each token of the source sentence with a text span that has a similar meaning but in the target style. The new span is generated via a \textit{non-autoregressive} masked language model, which can better preserve the local-contextual meaning of the replaced token. 
This RLM generation scheme gathers the flexibility of autoregressive models and the accuracy of non-autoregressive models, which bridges the gap between sentence-level and word-level style transfer methods.
 To control the generation style more precisely, we conduct a token-level style-content disentanglement on the hidden representations of RLM.
 Empirical results on real-world text datasets demonstrate the effectiveness of RLM compared with other TST baselines. The code is at \url{https://github.com/Linear95/RLM}.
\end{abstract}

\vspace{-2mm}
\section{Introduction}
\vspace{-2mm}
Text style transfer (TST) aims to rewrite natural language sentences with the same semantic meaning but in a different target style, where the styles of text are usually considered from the perspectives of formality~\citep{rao2018dear}, sentiment~\citep{hu2017toward,shen2017style}, persona~\citep{wu2021transferable}, and politeness~\citep{madaan2020politeness}, \textit{etc}.
Recently, text style transfer has increasingly attracted considerable interest for its potential usage in various scenarios such as personalized chatbots~\citep{wu2021transferable,zhang2018personalizing}, writing assistants~\citep{rao2018dear}, and non-player characters in games~\citep{campano2009socio,yunanto2019english}. 
Although with substantial application potential, TST remains a difficult learning task, mainly because of the scarcity of parallel source-target training data in practice~\citep{malmi2020unsupervised}. Therefore, current studies of TST have concentrated on the \textit{unsupervised} learning setup~\citep{huang2021nast,reid2021lewis,li2022text}, where only raw-text sentences and their corresponding source style labels are provided in training data.

Prior works of unsupervised TST fall into two tracks: \textit{sentence}-level and \textit{word}-level. Sentence-level  methods allow models to generate the entire transferred sentences. To control the generation, a large group of sentence-level methods disentangle source style information from the content representations in latent spaces, then generate the transferred sentences with the content representations and target style attributes~\citep{hu2017toward,lample2018multiple,john2019disentangled,cheng2020improving}. Other sentence-level methods design discriminators to evaluate and induce the generated sentence to reach the target style~\citep{yang2018unsupervised,holtzman2018learning}. Although remaining the mainstream, sentence-level methods have been continuously challenged with the \textit{content preservation} problem~\citep{huang2021nast}, that the transferred results always contain irrelevant words or differ from the semantic meaning of the source sentences, due to the overmuch freedom given to the neural generators.

To mitigate the content preservation problem, word-level TST methods keep the backbone of the original sentence, only masking or removing style-related words~\citep{li2018delete,sudhakar2019transforming,malmi2020unsupervised,huang2021nast}. Then the generators are supposed to fill words or spans in the target style into the masked or removed positions. With precise edits on source sentences, word-level methods have achieved success to retain the original content meaning in the generation. Besides, many of the word-level transfer methods conduct generation in non-autoregressive schemes~\citep{wu2019mask,huang2021nast,reid2021lewis}, which further improve the transfer efficiency compared with auto-regressive sentence-level methods~\citep{lample2018multiple,john2019disentangled}.
However, the performance of word-level methods highly depends on the selection of style-related words in source sentences. In fact, most of the detection methods of style-related words are man-made with heuristics~\citep{qian2021discovering}, which leads to large performance variance when applied in different transfer domains. In addition, this type of delete-and-refill methods lack generation diversity, for only a few words in the source sentence are changed after the transfer.



To take advantage of both sentence-level and word-level TST methods, we propose a novel sequence-to-sequence transfer framework called Replacing Language Model (RLM). Like sentence-level methods, our RLM autoregressively generates the transferred sentence in the target style, whereas the prediction of each target-sentence token is based on a word-level non-autoregressive masked language model. Besides, different from prior works disentangling text representations on sentence-level,  our RLM learns disentanlged style-content representations on  each token's latent embedding, further providing fine-grained control of the transfer process. Our model successfully collect the flexibility and the precision from sentence-level and word-level TST methods respectively. In the experimental part, we evaluate the effectiveness of our RLM on two real-world datasets (Yelp and Amazon) with respect to both automatic and man-made metrics, where our RLM outperforms other baselines in terms of the overall text style transfer quality. 
\vspace{-2mm}
\section{Background}
\vspace{-2mm}
\textbf{Language Modeling: }\label{sec:background-language-model}
Language Modeling (LM) is a fundamental task of natural language processing (NLP). Given a natural language sentence $\mX$ as a length-$n$ sequence of words $(\vx_0, \vx_1, \dots, \vx_{n-1})$, LM aims to learn the probability of the occurrence of $\mX$, \textit{i.e.}, $P(\mX) = P(\vx_0, \vx_1, \dots, \vx_{n-1})$. A well-learned language model can be useful in numerous NLP applications, such as text representation learning~\citep{kenton2019bert,liu2019roberta}, text generation~\citep{radford2019language,brown2020language}, and model performance evaluation~\citep{brown1992estimate,huang2021nast}. 
To learn the probability $P(\mX)$, a commonly used calculation paradigms~\citep{bengio2000neural} is:
\begin{align}
P(\mX) = P(\vx_0, \vx_1, \dots, \vx_{n-1})
  = P(\vx_0) \prod_{i=1}^{n-1} P(\vx_i| \vx_{0}, \dots, \vx_{i-1}) = \prod_{i=0}^{n-1} P(\vx_i | \mX_{0:i}), \label{eq:lm-autoregressive} 
\end{align}
where sub-sequence $\mX_{i:j} = (\vx_i, \vx_{i+1}, \dots, \vx_{j-1})$ if $i<j$, and $\mX_{i:j} = \text{\O}$ if $i \geq j$. With the decomposition in \eqref{eq:lm-autoregressive}, the model iteratively provides the probability of next-token prediction $\vx_i$, conditioned on 
 its prefix $\mX_{0:i}$. The calculation scheme in \eqref{eq:lm-autoregressive} is named the \textit{autoregressive} (AR) language modeling~\citep{brown2020language}. 

Although widely applied in NLP scenarios, AR models have been growingly challenged at their low-efficiency and accumulation errors~\citep{qi2021bang,arora2022exposure}. Hence, many non-autoregressive (NAR) modeling frameworks are recently proposed~\citep{kenton2019bert,shen2020blank}. Generally, NAR models have the following form:
\begin{equation}\label{eq:lm-NAR} 
  P(\mX) = P(\mX_{\gI}| \mX_{\bar{\gI}}) \cdot P(\mX_{\bar{\gI}}),
\end{equation}
where $\mX_{\gI} = \{ \vx_i \vert i\in \gI \}$ and $\mX_{\bar{\gI}} = \{\vx_i \vert i \notin \gI\}$ are selected from index set $\gI$ and its complement $\bar{\gI}$ respectively. Since the size of $\gI$ can vary, a practical implementation of \eqref{eq:lm-NAR} is masking words in $\mX_{\gI}$ with special \MASK tokens to keep the same input sequence length, then letting the language model reconstruct the masked sequence, which leads to masked language modeling (MLM)~\citep{kenton2019bert}.

\noindent \textbf{Disentangled Representations 
for Style Transfer:}\label{sec:DRL-MI-ST}
Disentangled representation learning (DRL) targets to map data instances into independent embedding sub-spaces, where different sub-spaces represent different attributes of the input data~\citep{locatello2019challenging}. By combining or switching the different subspace embedding parts, one can  obtain new latent codes with sufficient information of  the desired attributes. Then the operated latent codes can be further fed into a decoder to generate new instances with the target  attributes. With the effective control of generated attributes, DRL has been widely applied in style transfer tasks~\citep{lee2018diverse,john2019disentangled,yuan2020improving}.
Among DRL-based style transfer methods, \citet{yuan2020improving} learn disentangled style embedding $\vs$ and content embedding $\vc$  of each  data point $\vx$ from an information-theoretic perspective. More specifically, the mutual information (MI)~\citep{kullback1997information} $I(\vs;\vc)$ is utilized to measure the information overlap between style embedding $\vs$ and content embedding $\vc$:
\begin{equation}
    I(\vs; \vc) = \mathbb{E}_{P(\vs, \vc)} [ \log \frac{P(\vs, \vc)}{P(\vs) P(\vc)}],
\end{equation}
which computes the difference between joint distribution $P(\vs, \vc)$ and the product of marginal distributions $P(\vs) P(\vc)$. The learning objective is to minimize the MI between style embedding $\vc$ and content embedding  $\vc$, while maximize the MI  between combined latent $(\vs, \vc)$ and the observation $\vx$:
\begin{equation}\label{eq:DRL-MI-objective}
    \min I(\vs; \vc) - I(\vx; \vs, \vc).
\end{equation}
If the objective in \eqref{eq:DRL-MI-objective} is well-learned, one can replace the source style $\vs_\text{source}$ with the target style embedding $\vs_\text{target}$, and combine it with the source content $\vc_\text{source}$ as $(\vs_\text{source}, \vc_\text{target})$  without losing any content semantics. Then the joint $(\vs_\text{source}, \vc_\text{target})$ already contains sufficient style and content information to generate the transferred sentence $\vx_\text{transfer}$. However, calculating MI values are challenging without distribution closed-forms provided~\citep{poole2019variational}.  Details about the implementation of  \eqref{eq:DRL-MI-objective} with MI estimators are in Section~\ref{sec:RLM-learning-obj}.




\vspace{-2mm}
\section{Replacing Language Model}
\vspace{-2mm}
Given a sentence $\mX = ( \vx_0, \vx_1, \dots, \vx_{n-1})$ and a target style $\vs \in \gS$, text style transfer is to generate a corresponding $\mY = (\vy_0, \vy_1, \dots, \vy_{m-1})$, which has the similar content meaning but is rewritten in style $\vs$. We aim to introduce the  replacing language model (RLM)  to learn the probability $P(\mY | \mX, \vs)$ under the unsupervised setup, where only raw-text $\vx$ and its source style label $\bar{\vs} \in \gS$ are provided. 

The main idea of RLM is to iteratively replace each $\vx_i \in \mX$ with a new text span that has the same content meaning but in the target style $\vs$. By accumulating the generated new spans, we can obtain the transferred sentence $\mY$. To describe the RLM process in details, in Section~\ref{sec:RLM-equal-length}, we first discuss style transfer with a constrain that the target $\mY$ has the same length as the source $\mX$. The model design and learning loss of RLM are introduced in Section~\ref{sec:RLM-model-design} and Section~\ref{sec:RLM-learning-obj}, respectively.  Then we extend the framework into more general unequal-length transfer scenarios in Section~\ref{sec:RLM-inequal-length}.  


\vspace{-1.5mm}
\subsection{RLM for Equal-Length Transfer}\label{sec:RLM-equal-length}
\vspace{-1.5mm}
In this part, we first limit the transferred sentence $\mY$ to have the same length as the original input $\mX$, which is a simpler transfer scenario for discussion.
%
%
The target of RLM is to autoregressively (AR) replace each $\vx_i$ in $\mX$ with a corresponding $\vy_i$, such that $\vy_i$ has the same content meaning as $\vx_i$ but in target style $\vs$ ($\vy_i$ can equal $\vx_i$ if not style-related). We first present the sequence-to-sequence modeling of style transfer in the autoregressive paradigm:
\begin{equation}\label{eq:seq2seq-lm}
  P(\mY | \mX, \vs) = \prod_{i=0}^{n-1} P(\vy_i | \mX, \mY_{0:i}, \vs),
\end{equation}
where $\mY_{0:i} = (\vy_0, \vy_1, \dots, \vy_{i-1})$ is the generated prefix of $\vy_i$. Based on our objective of RLM, each $\vy_i$ and $\vx_i$ are supposed to have the same content information, so $\mY_{0:i}$ also contains (ideally) the equivalent content meaning as $\mX_{0:i}$ does.
%
We apply the Bayes' rule~\citep{box2011bayesian} on the token variable pair $(\vx_i,\vy_i)$ in each probabilistic term $P(\vy_{i} | \mX_{0:n}, \mY_{0:i}, \vs)$ of \eqref{eq:seq2seq-lm} and obtain:
\begin{align}
  P(\vy_{i} | \mX \mY_{0:i}, \vs) = P(\vy_{i} | \vx_i, \mX_{-i}, \mY_{0:i}, \vs) 
  = \frac{P(\vy_{i} | \mX_{-i} , \mY_{0:i}, \vs) \cdot
  P(\vx_{i}| \mX_{-i}, \mY_{0:i+1}, \vs)
  }{P(\vx_{i}| \mX_{-i} , \mY_{0:i}, \vs)} \label{eq:RLM-bayes-rule},
\end{align}
where $\mX_{-i} = \mX - \{\vx_i\} = \mX_{0:i} \cup \mX_{i+1:n}$ denotes sequence $\mX$ without token $\vx_i$. To further simplify, we analyze the three probabilistic terms on the right-hand side of \eqref{eq:RLM-bayes-rule}, respectively:
\begin{enumerate}[leftmargin=21pt, label=(\roman*)]
%
    \item For $P(\vy_{i} | \mX_{-i},  \mY_{0:i}, \vs)$, note that $\mY_{0:i}$ should have the equivalent content information as $\mX_{0:i}$. Besides, target style information $\vs$ is given.  One can select either $\mX_{0:i}$ or $\mY_{0:i}$ to provide sufficient prefix information (with style $\vs$) for target token $\vy_i$ prediction. Our choice is $\mY_{0:i}$, leading $P(\vy_{i} | \mX_{-i}, \mY_{0:i}, \vs) = P(\vy_{i} | \mY_{0:i},  \mX_{i+1: n}, \vs)$, for sequence $\mY_{0:i}$ offering more coherence than $\mX_{0:i}$ to generate the target $\vy_i$.
%
%
\item For $P(\vx_{i}| \mX_{-i}, \mY_{0:i+1}, \vs)$, $\mX_{0:i}$ includes the same content as $\mY_{0:i}$, and $\vs$ has the style information of $\mY_{0:i}$. Hence, we can use $(\mX_{0:i},\vs)$ to represents all the information from $\mY_{0:i}$, with only $\vy_i$ remained, $P(\vx_{i}| \mX_{-i}, \mY_{0:i+1}, \vs) = P(\vx_{i}| \mX_{0:i}, \vy_i, \mX_{i+1:n}, \vs)$.
Moreover, the original $\vx_i$ does not depend on the target style $\vs$, and $\mX_{-i}$ is supposed to have sufficient information of the source style.
Therefore, we further use $ P(\vx_{i}| \mX_{0:i}, \vy_i, \mX_{i+1:n})$ to approximate
$P(\vx_{i}| \mX_{0:i}, \vy_i, \mX_{i+1:n}, \vs)$. 

\item For the denominator $P(\vx_{i}| \mX_{-i}, \mY_{0:i}, \vs)$, similarly to analysis in (ii), we can remove condition $\mY_{0:i}$, so that 
$  P(\vx_{i}| \mX_{-i} , \mY_{0:i}, \vs) = P(\vx_{i}| \mX_{-i}, \vs)$.
Furthermore,  $P(\vx_{i}| \mX_{-i}, \vs)$ is not related to any model output $\vy_i$, which can be treated as a constant to model parameters.
%
\end{enumerate}
Based on the above discussion, we can simplify the right-hand side of \eqref{eq:RLM-bayes-rule} as follows:
\begin{align}
   P(\vy_{i}  | \mX, \mY_{0:i}, \vs)   \propto \label{eq:simplified-RLM} \underbrace{P(\vy_{i} | \mY_{0:i}, \mX_{i+1: n}, \vs)}_\text{Prediction Term} \cdot \underbrace{P(\vx_{i}| \mX_{0:i}, \vy_i, \mX_{i+1:n})}_\text{Reconstruction Term}.
\end{align}
The prediction term $P(\vy_{i} | \mY_{0:i}, \mX_{i+1: n}, \vs)$ provides the $i$-th token $y_i$ in the target sentence based on the generated $\mY_{0:i}$ and the remained $\mX_{i+1:n}$. While the reconstruction term $P(\vx_{i}| \mX_{0:i}, \vy_i, \mX_{i+1:n})$ gives the probability to reconstruct original $\vx_i$ with the new $\vy_i$ inserted back to the $i$-th position of the source sentence.

Intuitively, prediction term $P(\vy_{i} | \mY_{0:i}, \mX_{i+1: n}, \vs)$ induces  $\vy_i$ to be coherent with the prefix $\mY_{0:i}$ in target style $\vs$, and consistent with the contextual information of $(\mY_{0:i}, \mX_{i+1:n})$. Different from directly predicting $P(\vy_i | \mX, \mY_{0:i}, \vs)$~\citep{lample2018multiple,lai2021thank}, term $P(\vy_{i} | \mY_{0:i}, \mX_{i+1: n}, \vs) $ utilizes less but sufficient semantic information from the source sentence, with the position $i$ highlighted for a more accurate token-level generation.
Meanwhile, $P(\vx_i| \mX_{0:i}, \vy_i, \mX_{i+1:n})$ measures the content information $\vy_i$ carrying from original $\vx_i$, by letting $\vy_i$ reconstruct $\vx_i$ with contexts $\mX_{0:i}$ and $\mX_{i+1:n}$.
Consequently, we obtain \textbf{R}eplacing \textbf{L}anguage \textbf{M}odel (RLM) for equal-length transfer:
\begin{equation}
 P(\mY | \mX ,\vs) \propto  P_{\text{RLM}} (\mY | \mX, \vs)= 
  \prod_{i=0}^{n-1} P(\vy_{i} | \mY_{0:i}, \mX_{i+1: n}, \vs) \cdot P(\vx_{i}| \mX_{0:i}, \vy_i, \mX_{i+1:n}). 
\end{equation}
Practically in the $i$-th generation step, we select $\vy^*_i$ that maximizes $P(\vy_i|\mX, \mY^*_{0:i}, s) $ as the $i$-th prediction. Within our RLM scheme, we can predict $\vy^*_i$ from:
\begin{equation}\label{eq:topk-selection}
      \max_{\vy_i \in \gY}  P(\vy_{i} | \mY^*_{0:i}, \mX_{i+1: n}, \vs) \cdot {P(\vx_{i}| \mX_{0:i}, \vy_i, \mX_{i+1:n})}.
\end{equation}
Theoretically, the candidate set $\gY$ can have the same size as the whole vocabulary $\gV$. To reduce the computational complexity, we approximate the candidates in  $\gY$ with tokens $\vy_i$ that have the top-$K$ logits of term $P(\vy_i | \mY_{0:i}, \mX_{i+1:n}, \vs)$ as $\gY_i^K= \{\vy\in\gV| \text{ value }  P(\vy | \mY_{0:i}, \mX_{i+1:n}, \vs) \text{ is top-}K\}$.

\vspace{-1.5mm}
\subsection{Model Design}\label{sec:RLM-model-design}
\begin{figure*}[t]
  \centering
  \includegraphics[width=0.98\textwidth]{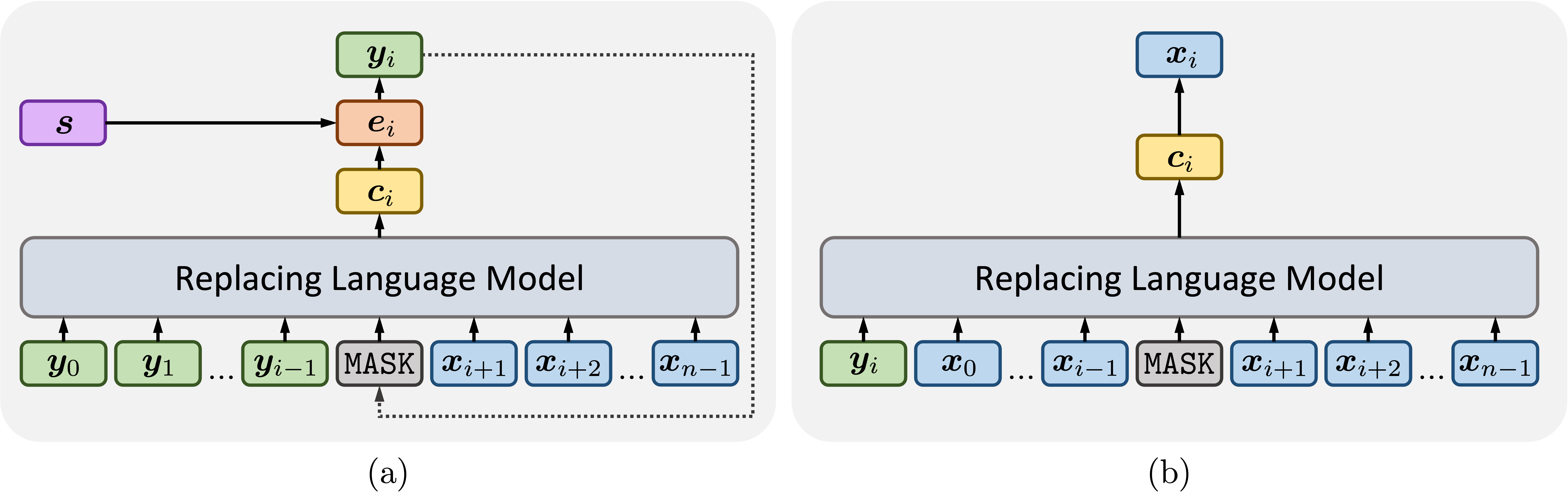}
  \vspace{-2mm}
  \caption{Replacing language model (RLM) for equal-length transfer. (a)~Prediction term: The generated $\mY_{0:i}$ and original $\mX_{i+1: n}$ are combined with a \MASK token and fed into the transformer-based encoder. RLM outputs content embedding $\vc_i$ at \MASK position, and fuse it with target style embedding $\vs$ to predict the $i$-th token $\vy_i$, providing $P(\vy_i | \mY_{0:i}, \mX_{i+1:n}, \vs)$. Then $\vy_i$ is inserted back at \MASK position and $\vx_{i+1}$ will be masked for the $(i+1)$-th generation setup. (b)~Reconstruction term: The prediction candidate $\vy_i$ is set in front of masked sequence $\mX_{-i}$. RLM reconstructs original $\vx_i$ based on content $\vc_i$ at \MASK position with the probability $p(\vx_i| \mX_{0:i}, \vy_i, \mX_{i+1:n})$. }
\label{fig:RLM-framework-equal-length}
\vspace{-3.5mm}
\end{figure*}

\vspace{-1.5mm}
 Next, we describe how to parameterize the prediction term $P(\vy_{i} | \mY_{0:i}, \mX_{i+1: n}, \vs)$ and the reconstruction term $P(\vx_{i}| \mX_{0:i}, \vy_i, \mX_{i+1:n})$ in \eqref{eq:simplified-RLM} with neural networks. Both terms generate one token at the $i$-th position of the input sequence, offering us the convenience to parameterize both of them with similar model structures.
As in Figure~\ref{fig:RLM-framework-equal-length}, we build the RLM encoder $E_{\text{RLM}}(\cdot)$ based on a BERT~\citep{kenton2019bert} encoder pretrained on masked language modeling tasks.
%
%
%

\begin{figure}[t]
    \centering
\includegraphics[width=0.86\textwidth]{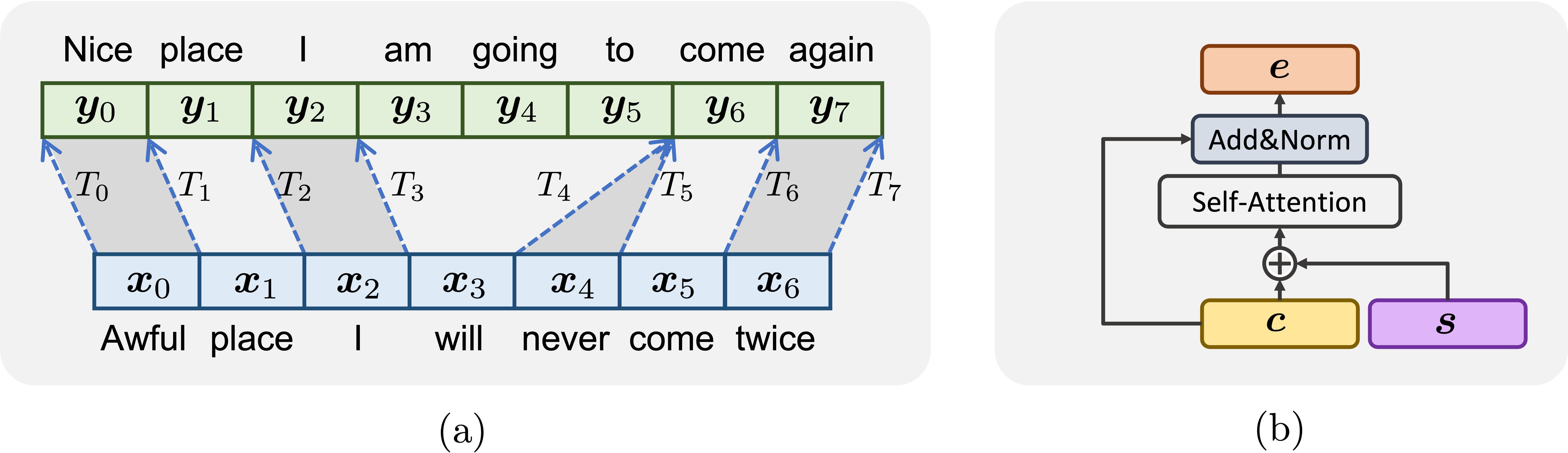}
   \vspace{-2.8mm}
    \caption{(a) An example of sentence alignment. The blue sequence is the source and the green one is the target. Each $T_i$ points token $\vx_i$ to the start position of the same-content transferred text span. (b) Fusion of style and content embeddings. The target style $\vs$ is concatenated to every content $\vc$ of input tokens. Then a self-attention is conducted on the combined embedding sequence. The attention results are added to the original content embedding with layer normalization.}
    \label{fig:align-and-res}
   \vspace{-4.mm} 
\end{figure}

In Figure~\ref{fig:RLM-framework-equal-length}(a),  we insert a \MASK token between the generated $\mY_{0:i}$ and the remained $\mX_{i+1:n}$ , then feed this mixed sequence through the RLM encoder $E_\text{RLM}(\cdot)$. The output at the masked position  is the content embedding $ \vc_i = E_\text{RLM}(\mY_{0:i}, \texttt{[MASK]}, \mX_{i+1:n})$,
%
%
where $\vc_i$ is supposed to be filtered with any input style information and only contains the content information of the masked input position.
 Next we combine $\vc_i$ with the target style embedding $\vs$ to obtain a overall representation $\ve_i = f(\vs, \vc_i)$ (as shown in Figure~\ref{fig:align-and-res}(b)), and make the $\vy_i$ prediction with a prediction head $H_\text{pred}(\cdot)$, 
 \begin{align}
 P_\text{RLM}(\vy_i| \vs, \vc_i) = \text{Softmax}(H_\text{pred}(f(\vs, \vc_i)) 
 =P_{\text{RLM}} (\vy_i | \mY_{0:i}, \mX_{i+1:n}, \vs). \label{eq:style-content-prediction}
 \end{align}
In Figure~\ref{fig:RLM-framework-equal-length}(b), we reconstruct $\vx_i$ based on $\mX_{-i}$ and $\vy_i$ also within a masked language modeling scheme. We do not   directly place $\vy_i$ at the $i$-th token position between $\mX_{0:i}$ and $\mX_{i+1:n}$. Instead, we set $\vy_i$ at the start of the sentences as a prompt~\citep{gao2021making}. Practically we find that putting $\vy_i$ at the $i$-th position always results in the same output as the model prediction, which is probably because pretrained MLM transformers learn over-strong hidden representation. 
Therefore, we obtain the content embedding at the masked position as,  
\begin{equation}\label{eq:encode-content-reconstruct}
\vc'_i = E_\text{RLM}(\vy_i, \mX_{0:i}, \texttt{[MASK]}, \mX_{i+1:n}).  
\end{equation}
  This time we directly reconstruct $\vx_i$ from content $\vc'_i$ via another prediction head $H'_\text{pred}(\cdot)$:
\begin{align}
    P_\text{RLM}(\vx_i | \vc'_i) = \text{Softmax}(H'_\text{pred}(\vc'_i))
    =P_{\text{RLM}}(\vx_i | \mX_{0:i}, \vy_{i}, \mX_{i+1:n}). \label{eq:decode-content-only}
\end{align}
%
%

\vspace{-1.5mm}
\subsection{Learning Objective}\label{sec:RLM-learning-obj}
\vspace{-1.5mm}
 To learn the parameterized terms $P_{\text{RLM}} (\vy_i | \mY_{0:i}, \mX_{i+1:n}, \vs)$ and $ P_{\text{RLM}}(\vx_i | \mX_{0:i}, \vy_{i}, \mX_{i+1:n})$,  we can only access raw-text sentences $\mX$ and their source label $\bar{\vs}$  under the unsupervised setup, without paralleled target $\mY$ provided. Based on our model design, the RLM encoder $E_\text{RLM}(\cdot)$ only encodes the content information of the masked position. Therefore, we can  encode the source sentence to obtain the content embedding instead, which is:
\begin{align}
    \vc_i =E_{\text{RLM}}(\mY_{0:i}, \texttt{[MASK]}, \mX_{i+1:n})
    =&E_{\text{RLM}}(\mX_{0:i}, \texttt{[MASK]}, \mX_{i+1:n}), 
    \label{eq:encode-masked-sentence-1} \\
    \vc'_i = E_\text{RLM}(\vy_i, \mX_{0:i}, \texttt{[MASK]}, \mX_{i+1:n}) 
    =& E_\text{RLM}(\vx_i, \mX_{0:i}, \texttt{[MASK]}, \mX_{i+1:n}). 
    \label{eq:encode-masked-sentence-2}
\end{align}
%
%


For the prediction term $P_{\text{RLM}} (\vy_i | \mY_{0:i}, \mX_{i+1:n}, \vs)$,   inspired by \citep{yuan2020improving}, we design our learning loss from an information-theoretic perspective to eliminate style information from the content embeddings. We minimize the mutual information (MI) $I(\bar{\vs};\vc_i)$ to reduce the information overlap of $\bar{\vs}$ and $\vc_i$. Meanwhile, $\bar{\vs}$ and $\vc_i$ should have rich semantic information from the input token $\vx_i$, so we maximize $I(\vx_i; \bar{\vs}, \vc_i)$. 
   The overall learning objective is:
\begin{equation}
\label{eq:learning-obj}
    \min I(\bar{\vs};  \vc_i) - I(\vx_i; \bar{\vs}, \vc_i). 
\end{equation}
Practically, the MI values are difficult to calculate without knowing the distribution closed-forms~\citep{poole2019variational}, so we utilized two sample-based MI bounds to estimate. 
%
%
 To maximize  $I(\vx; \bar{\vs}, \vc_i)$, we maximize a variational lower bound~\citep{agakov2004algorithm} instead:
\begin{align}
I(\vx; \bar{\vs}, \vc_i) \geq  \mathbb{E}_{P(\vx_i, \bar{\vs}, \vc_i)} [\log P_\text{RLM}(\vx_i| \bar{\vs}, \vc_i) ] - \mathbb{E}_{P(\vx_i)}[\log P(\vx_i)],
\end{align}
where $P_\text{RLM}(\vx| \vs, \vc)$ is the prediction head defined in \eqref{eq:style-content-prediction},
and $\mathbb{E}_{P(\vx_i)}[\log P(\vx_i)]$ is a constant based on input $\vx_i$. Therefore, we can maximize the  log-likelihood $\mathbb{E}[\log P_\text{RLM}(\vx_i | \bar{\vs}, \vc_i)]$ for $I(\vx; \bar{\vs}, \vc_i)$ maximization. To minimize $I(\bar{\vs}; \vc_i)$, we follow \citet{cheng2020club} and minimize a variational MI upper bounds:
\begin{align}
I(\bar{\vs} ; \vc_i) \leq \mathbb{E}_{P(\bar{\vs}, \vc_i)}[\log  Q(\bar{\vs}| \vc_i)] - \mathbb{E}_{P(\bar{\vs})P(\vc_i)} [\log Q(\bar{\vs}| \vc_i)], \nonumber
\end{align}
where $Q(\bar{\vs} | \vc_i)$ is an variational approximation to the conditional distribution $P(\bar{\vs} | \vc_i)$.

\begin{figure*}[t]
    \centering
    \includegraphics[width=0.98\textwidth]{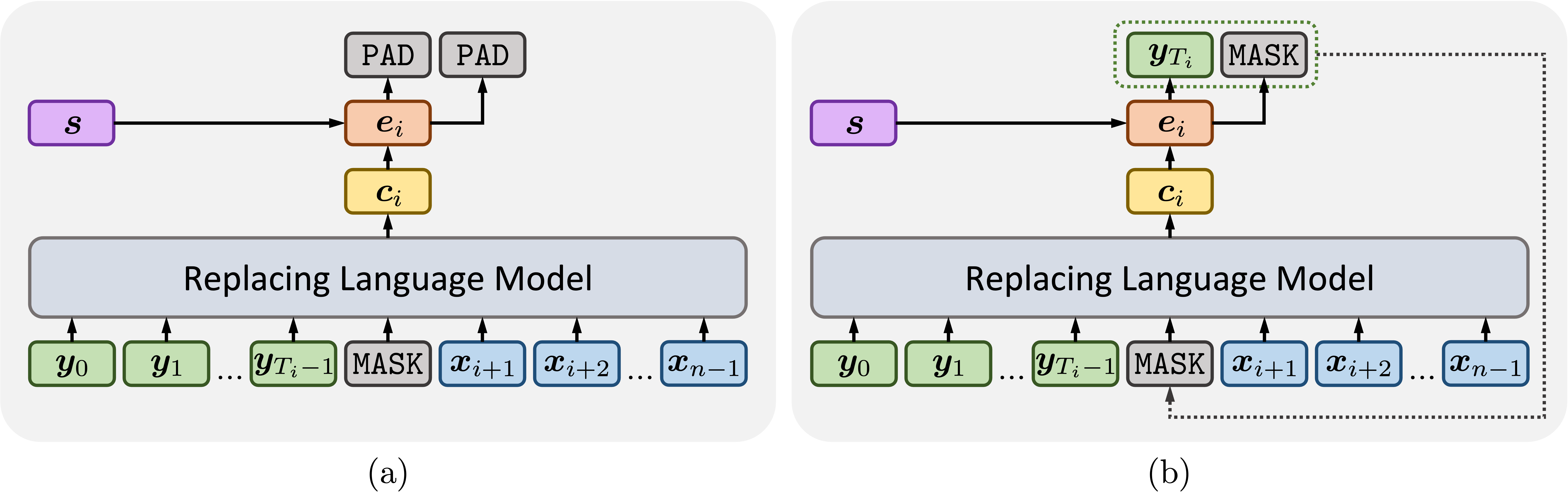}
    \vspace{-2.8mm}
    \caption{Unequal-length transfer. (a)~Deletion: if token $\vx_i$ is supposed to be deleted in the target sentence, RLM will output a \PAD token at the \MASK position. (b)~Insertion: to insert a token in the target sentence, the next-token prediction head will output a \MASK token. Then the generated $\vy_{T_i}$ and new \MASK tokens are inserted back into the input, and the next-step masked language model generation will be conducted on the generated \MASK token, instead of masking $\vx_{i+1}$.} 
    \label{fig:inequal-transfer-adapation}
\end{figure*}

To learn $ P_{\text{RLM}}(\vx_i | \mX_{0:i}, \vy_{i}, \mX_{i+1:n})$, we collect the candidates $\vy_i$ with the top-$K$ logits of $P_{\text{RLM}} (\vy_i | \mY_{0:i}, \mX_{i+1:n}, \vs)$, and replace $\vx_i$ with each of them as in \eqref{eq:encode-masked-sentence-2}, then maximize the probability of  $P_\text{RLM}(\vx_i | \vc'_i)$ to reconstruct the original $\vx_i$. More details about learning objective are provided in the Supplementary Materials.

Unlike prior TST methods attempting to learn a sentence-level disentangled content vector and transfer the whole sentence with it, we disentangle style and content information at word-level, making it easier for deep networks to fit and infer.
Moreover, our learning objective limits the generation at the masked positions, providing more fine-grained guidance to the transfer model. 

\vspace{-1.5mm}
\subsection{Adaptions to Unequal-Length Transfer}\label{sec:RLM-inequal-length}
\vspace{-1.5mm}

In more general scenarios, transferred sentence $\mY$ usually has different lengths than the original sentence. For unequal-length transfer, we relax the objective of RLM, so that $\vx_i$ 
can be replaced by a variable-length text span $\mY_{T_i:T_{i+1}}$, where alignment $T_i$ is the starting index of the $i$-th generated span in $\mY$, $T_i \leq T_{i+1}$, as shown in Figure~\ref{fig:align-and-res}(a). Similarly to Section~\ref{sec:RLM-equal-length}, we can split the sequence $\mY$ at each position $T_i$, and apply the Bayes' Rule to each token $\vx_i$ and its corresponding $\mY_{T_i: T_{i+1}}$ to obtain the general RLM form:
\begin{align}
 P_{\text{RLM}} (\mY | \mX, \vs)=  \prod_{i=0}^{n-1} \Big[  P(\mY_{T_i:T_{i+1}} | \mY_{0:T_i}, \mX_{i+1: n}, \vs)
 \cdot P(\vx_{i}| \mX_{0:i}, \mY_{T_i:T_{i+1}}, \mX_{i:n})\Big]. \label{eq:RLM-unequal-length}
\end{align}
We make some modifications to the equal-length RLM framework. For reconstruction term $P(\vx_{i}| \mX_{0:i}, \mY_{T_i:T_{i+1}}, \mX_{i:n})$, we change the input sequence to $(\mY_{T_i:T_{i+1}}, \mX_{0:i}, \texttt{[MASK]}, \mX_{i+1:n})$ in \eqref{eq:encode-content-reconstruct}, then the output logits at \MASK still provides $P_{\text{RLM}}(\vx_{i}| \mX_{0:i}, \mY_{T_i:T_{i+1}}, \mX_{i+1:n})$. For generation term $P(\mY_{T_i:T_{i+1}} | \mY_{0:T_i}, \mX_{i+1: n}, \vs)$, we consider three different situations:
\begin{enumerate}[leftmargin=21pt, label=(\roman*)]  \item If $T_{i+1}- T_i =0$, which means $\mY_{T_i:T_{i+1}} = \text{\O}$, we let the prediction head $H_\text{pred}$ output a special \PAD token at the \MASK input position, as shown in Figure~\ref{fig:inequal-transfer-adapation}(a).
  \item If $T_{i+1} - T_i = 1$, which means only one token $\vy_{T_i}$ is in the text span $\mY_{T_i:T_{i+1}}$ for replacing $\vx_i$, the situation is exactly the same as equal-length RLM in Figure~\ref{fig:RLM-framework-equal-length}(a).
  \item If $T_{i+1}- T_{i} > 1$, besides predicting $\vy_{T_i}$, we should insert more token into the span $\mY_{T_i:T_{i+1}}$. We introduce an insertion head $H_\text{insert}(\cdot)$ which takes the fused embedding $\ve_i$ and outputs a binary prediction of \MASK or \PAD. As shown in Figure~\ref{fig:inequal-transfer-adapation}(b), besides the prediction head $H_\text{pred}$ providing $\vy_{T_i}$, the insertion head $H_\text{insert}$ will output a \MASK token. In the next generation step, RLM will generate with the predicted \MASK from $H_\text{insert}$, instead of masking $\vx_{i+1}$. This insertion process continues until $H_\text{insert}$ outputs a \PAD token.
\end{enumerate}
To train  $H_\text{insert}$, we can remove more than one token from the input sequence $\mX$, and let $H_\text{insert}$ predict whether new tokens should be inserted. For input $(\mX_{0:i}, \texttt{[MASK]}, \mX_{i+k:n})$, if $k>=2$, $H_\text{insert}$ has to predict a \MASK token, otherwise $H_\text{insert}$ should output a \PAD token.

\vspace{-2mm}
 \section{Related Work}
 \vspace{-2mm}
\noindent\textbf{Sentence-level}: Sentence-level methods generate the entire sentence with neural network based generators. Among sentence-level TST, \citet{shen2017style} learn a shared content embedding space across sentences with  different styles, in which style information is eliminated. Then the neural generator combines the disentangled content embedding and target style to predict the transferred sentence. \citet{hu2017toward} build a variational auto-encoding~\citep{kingma2013auto} framework to encode the disentangled content embedding, and further adds style discriminators to induce the generator to output sentences with the target style. Different from the disentangling methods, \citet{dai2019style} propose a style transformer trained with cycle-consistency losses, and conducts direct transfer with source sentence and target style fed together into a transformer model. \citet{he2019probabilistic} treat the target sentence as a sequence of latent variables of the observed source sentence, then applies the variational inference to learn the latent-observation mapping.

\noindent\textbf{Word-level}: Word-level TST methods make fine-grained edits on the original source sentence, where style-related words are usually removed, or replaced by other words in the target style.
\citet{li2018delete} detect style-related words based on the word occurrence frequencies in corpus with different styles, then deletes high correlation words to retrieve a new sentence in the target style. \citet{sudhakar2019transforming} extend the delete-retrieve-generate method~\citep{li2018delete} with transformers, where keyword prompts are used to further improve content preservation. \citet{wu2019mask} utilize the pretrained masked language model to generate target-related words at the masked positions, in which source-related words are originally located. \citet{malmi2020unsupervised} extend mask-and-refill which enables variable-length generation at the masked positions. \citet{li2022text} iteratively replace a source style related span with a generated target-style span, until the pretrained text classifier outputs a high probability on the target style label. All the mentioned word-level TST methods have heuristic pre-processes to detect style-related words.




\vspace{-2mm}
\section{Experiments}
\vspace{-2mm}
We first describe the implementation details of the RLM model. Next, the three commonly-used automatic evaluation metrics and human evaluation are introduced to test the transfer performance. Then we compare our RLM with other baselines and further discuss the impact of hyper-parameters in the ablation study.

\vspace{-1.5mm}
\subsection{Implementation Details}\label{sec:implement-details-RLM}
\vspace{-1.5mm}


\textbf{Model Details} We construct our RLM based on pretrained transformer-based sentence encoder BERT~\citep{kenton2019bert}. More specifically, the pretrained BERT takes the masked sentence sequence (described in \eqref{eq:encode-masked-sentence-1}) as input, and output the last hidden state at the masked position as $\vh_i$. We use an attention block to extract only the content information $\vc_i$ from $\vh_i$. 
The style embeddings are set to learnable vectors and  initialized by  style description word embeddings from the  pretrained language models.
The fused embedding $\ve_i$ is output by a residual block~\citep{he2016deep} as shown in Figure~\ref{fig:align-and-res}(b). Instead of simply combining $\vs$ and $\vc_i$, we consider the contextual influence of $\vs$ on $\vc_i$. More specifically, we concatenate $\vs$ to every content embedding of the input sequences, then process a single-head self-attention on all concatenated embeddings. Then we add the attention results back to the content embedding sequence with layer normalization.

Follow the setting in \citet{kenton2019bert} and \citet{liu2019roberta}, the style and content embedding dimension is set to 768, which equals the dimension of transformer hidden states.
The prediction head $H_\text{pred}$ is initial with the pretrained masked language modeling head from BERT. 
The insertion head $H_\text{insert}$ is a pretrained pooling module from BERT following by a single fully connected layer, outputting the probability of next-token insertion. 

\noindent \textbf{Data Preparation}  We conduct the experiment on two real-world sentiment transfer datasets:  Yelp and Amazon Review. 
Yelp contains 450K restaurant reviews and business reviews.  Each review is labeled as positive or negative. Amazon review contains 550K product reviews and each is labeled as positive or negative similar to Yelp. We follow the data pre-processing setup from \citep{li2018delete} for a fair comparison.
To prepare the training data for masked language model fine-tuning,
for each training sample, we randomly select one word for masking, then reform the input sequence as in \eqref{eq:encode-masked-sentence-1} and \eqref{eq:encode-masked-sentence-2}. To enhance the quality of the masked training samples, when masking we skip all numbers and the pronouns defined in NLTK~\citep{bird2006nltk}, which are high-frequency words with no style-related information. Apart from that, we also skip words that are less style-related by following the design of attribute markers from ~\citep{li2018delete}. Samples within the same batch are of similar lengths to avoid noise in mask prediction brought by unnecessary lengthy paddings. 




\noindent \textbf{Training Setups}  We finetune the pretrained  BERT base model with AdamW~\citep{loshchilov2018decoupled} as the training optimizer,  with the learning rate set to $5\times 10^{-5}$.  The batch size is set to 16. 
Other parameters follow the initial settings of the pre-trained base BERT.

\vspace{-1.5mm}
\subsection{Evaluation Metrics} 
\vspace{-1.5mm}


\textbf{Style Transfer Accuracy}:   Each transferred sentence is sent into a pretrained style classifier to test whether it can be recognized correctly with the target style. The classification accuracy (ACC) is reported. This is evaluated by a RoBerTa ~\citep{Roberta} classifier which achieves an accuracy of 97.2\% on Yelp and 86.1\% on Amazon.

     \textbf{Content Preservation}: We use two metrics to measure the transfer content preservation: (1) Self-BLEU (S-BLEU)~\citep{papineni2002bleu}  is calculated between the generated sentence and the input source sentence, providing a rough criterion of content preservation; (2) Ref-BLEU (R-BLEU) is calculated between the generated sentence and man-made transfer results provided on the testing set, which reflects the similarity between the model's transfer and human rewriting. 
     
     
     \textbf{Overall Quality}: Following \citet{lee2021enhancing}, we take the geometric mean (GM) of the above automatic evaluation metrics as an overall transfer quality score.
     
       \textbf{Human Evaluation}: We also ask  annotators to score 1-5 to the transferred sentences, from the perspectives  of style transfer accuracy (SA), content preservation (CP) and sentence fluency (SF). The overall quality is also computed as the geometric mean (GM) of SA, CP, and SF.

\vspace{-1.5mm}
\subsection{Performance}
\vspace{-1.5mm}
We compared our replacing language model with several competitive baselines: CrossAlign~\citep{shen2017style}, DRG~\citep{li2018delete}, Style-Transformer~\citep{dai2019style}, CPVAE~\citep{xu2020betaVAE}, and RACoLN~\citep{lee2021enhancing}. 
The left-hand side of  Table~\ref{tab:all-evaluation-yelp} and Table~\ref{tab:automatic-amazon-evaluation} show the automatic evaluation results on both Yelp and Amazon datasets, respectively. Besides, Table~\ref{tab:automatic-amazon-evaluation} demonstrates the automatic evaluation on Amazon Review.

\begin{table*}[t]
  \centering
  \caption{Automatic and human evaluation of transfer results on the Yelp datasets.}
  \vspace{-2.5mm}
  \resizebox{0.95\textwidth}{!}{
     \begin{tabular}{l|rrr|r|rrr|r}
        \toprule[1.2pt]
        & \multicolumn{4}{c|}{\textbf{Automatic Evaluation}} & \multicolumn{4}{c}{\textbf{Human Evaluation}}\\
        
    \hline
          & \textbf{ACC} & \textbf{R-BLEU} & \textbf{S-BLEU} & \textbf{GM} &   \bf SA & \bf CP & \bf SF & \bf GM \\
          \hline
   CrossAlign~\citep{shen2017style} & 74.2 & 4.2 & 13.2 & 16.0   & 3.4 & 2.1 & 4.1 & 3.1  \\
 DRG~\citep{li2018delete} & 88.3 & 23.1 & 44.4 & 44.9  & 3.5 & 2.9 & 4.2 & 3.5\\
S-Transformer~\citep{dai2019style} & 87.3 & 19.8 & 55.2& 45.7 & 4.0 & \textbf{3.8} & 4.3 & 4.0 \\
 CPVAE~\citep{xu2020betaVAE} & 55.4 & 26.4 & 48.4 & 41.4 & 3.5 & 2.7 & \textbf{4.4} & 3.5  \\
RACoLN~\citep{lee2021enhancing} & \textbf{91.3} & 20.0 & \textbf{59.4}& 47.7  & 4.1 & {3.7} & 4.2 & 4.0 \\
\hline
RLM (Ours) & 91.0 & \textbf{30.6} & 51.7 & \textbf{52.4}  & \textbf{4.5} & {3.7} & 4.3 & \textbf{4.2} \\
  \bottomrule[1.2pt]
    \end{tabular}
   }
  \label{tab:all-evaluation-yelp}%
  \vspace{-3.5mm}
\end{table*}%


For automatic evaluation, on the Yelp dataset, our RLM reaches the highest Ref-BLEU score with a significant gap with baseline methods, which indicates our RLM's generation ability to transfer sentences with higher naturalness and closer to the human-made references. For style transfer accuracy, our RLM is competitive with the state-of-the-art baseline RACoLN~\citep{lee2021enhancing}. Although not conspicuous enough on the Self-BLEU metric,  our RLM still outperforms previous methods distinctly with respect to the overall performance (the geometric mean of ACC, R-BLEU, and S-BLEU). On the Amazon review dataset, CrossAlign~\citep{shen2017style} reaches the highest style transfer accuracy, but performs poorly under the Ref-BLEU and Self-BLEU measurements. In contrast, RACoLN~\citep{lee2021enhancing} achieves the best Ref-BLEU and Self-BLEU scores. However, the style transfer accuracy of RACoLN is apparently at a disadvantage. Although our RLM does not reach the highest score on any of the three metrics, the overall performance (GM) score is the highest among all the methods, indicating RLM with a better style-content trade-off. 

On the right-hand side of Table~\ref{tab:all-evaluation-yelp}, we report the human evaluation results of sentiment transfer on the Yelp review dataset. From the results, our RLM reaches the best style transfer accuracy (SA) with an average score 4.5, which distinctly outperforms other baselines. Besides, the content preservation (CP) score (with average 3.7) and the sentence fluency (SF) score (with average 4.3) are competitive to the best records among all the baselines (CP average 3.8 and SF average 4.4). The geometric mean (GM) shows the overall transfer performance under the human judgement, where our RLM also surpasses other baselines evidently.

\begin{table}
  \centering
  \caption{Automatic evaluation on the Amazon dataset.}
 \vspace{-2.mm}
  \resizebox{0.55\textwidth}{!}{
    \begin{tabular}{l|rrr|r}
    \toprule[1.2pt]
          & \textbf{ACC} & \textbf{R-BLEU} & \textbf{S-BLEU} & \textbf{GM} \\
    \hline
CrossAlign          & \textbf{65.0} & 9.2 & 20.7 & 23.1  \\
 S-Transformer     & 58.3 & 27.7 & 57.3 & 45.2 \\
 CPVAE              & 40.0 & 28.6 & 39.7 & 35.7  \\
 RACoLN             & 48.7 & \textbf{36.1} & 54.5 & 45.7  \\
\hline
  RLM          & 57.5 & 30.9 & \textbf{54.7} & \textbf{46.0} \\
    \bottomrule[1.2pt]
    \end{tabular}
    }
    \vspace{-3.5mm}
  \label{tab:automatic-amazon-evaluation}%
\end{table}%

Based on the reported results above, we find that our RLM consistently remain a better overall transfer performance compared to other methods,  supporting the claim that our RLM has a comprehensive transfer ability with more balanced rewriting results.


\vspace{-1.5mm}
\subsection{Ablation Study}
\vspace{-1.5mm}

For the ablation study, we first consider the impact of the hyper-parameter $K$ in the top-$K$ selection of \eqref{eq:topk-selection}. We choose different $K$ values, and report the evaluation results in Table~\ref{tab:ablation-study}. From the results, we discover a clear trade-off between the style transfer accuracy (ACC) and the self-BLEU score. As the hyper-parameter $K$ goes larger, the style transfer accuracy increases. Because the prediction candidate set inflates, the RLM model has more flexibility to select word candidates to fit the target style. At the same time, the self-BLEU score goes down since the more prediction candidates, the lower probability for RLM to select the original input tokens. 
\begin{table}[h]
\vspace{-1.5mm}
  \centering
  \caption{Ablation study of RLM on the Yelp dataset.}
  \vspace{-2.5mm}
\resizebox{0.59\textwidth}{!}{
    \begin{tabular}{l|rrr|r}
    \toprule[1.2pt]
     RLM     & \bf ACC & \bf R-BLEU & \bf S-BLEU & \bf GM \\
    \hline
 Top-1 &  89.7 & 30.0 & \textbf{53.5} & \textbf{52.4}\\
Top-3 & 90.7 & 29.5 & 51.8 & 51.8\\
Top-5 &\textbf{ 91.0} & 30.6 & 51.7 & \textbf{52.4}\\
Top-10 & \textbf{91.0} & 29.5 & 51.5 & 51.7 \\
\hline
No $I(\vs;\vc)$ & 74.6 & 27.7 & 38.2 & 42.9 \\
No Insert & 89.6 & \textbf{31.0} & 50.1 & 51.8 \\
No Delete & 85.1 & 29.7 & 49.9 & 50.1 \\
No Delete\&Insert & 88.7 & 29.1 & 53.3 & 51.6 \\
    \bottomrule[1.2pt]
    \end{tabular}
}
\vspace{-2.5mm}
  \label{tab:ablation-study}%
\end{table}%

Besides, we test the effectiveness of different parts in RLM. We fix model with $K=5$ and train it without the disentangling loss $I(\vs;\vc)$. From Table~\ref{tab:ablation-study}, without the disentangling term $I(\vs;\vc)$, the performance goes much worse, which means the model cannot learn disentangled style and content representations without the mutual information minimization process. Also, we remove the insertion and deletion mechanism,  leading to slightly lower transfer performance, because the sentiment transfer on Yelp does not require large amount of text reorganizing. In addition, the reference sentences have heavy overlaps with the source sentences, leaving little space for inserting and deleting operation.

\vspace{-2mm}
\section{Conclusion}
\vspace{-2mm}
This paper introduces a new sequence-to-sequence  framework for text style transfer called replacing language model (RLM). In virtue of pretrained language models such as BERT, our model autoregressively predicts a target text span  based on the generated  prefix and the remaining source suffix in the masked language modeling (MLM) paradigm, and further scores the newly generated span based on the probability of reconstructing the corresponding word in the original sentence. Moreover, unlike prior \textit{sentence}-level disentangling methods, we eliminate the style information from the \textit{word}-level content embeddings with information-theoretic guidance, providing fine-grained control to generate transferred tokens. The empirical results on  Yelp and Amazon review dataset demonstrate the effectiveness of the proposed RLM. As a novel generation scheme, RLM combines autoregressive generators' flexibility and non-autoregressive models' accuracy. However, a potential limitation is about the transfer diversity, that RLM does not rewrite sentences with different word orders (\text{e.g.}, active voice to passive voice), which might be an advantage in contrast to other sequence-sequence tasks requiring order preservation, such as voice conversion.
From this perspective, We believe the replacing language model will have a further impact in broader sequence-to-sequence modeling scenarios such as machine translation, text rewriting, and speech processing.

\bibliography{reference}
\bibliographystyle{plainnat}

\clearpage
\onecolumn
\appendix

\section{Transferred Samples}
\begin{table}[htbp]
  \centering
    \begin{tabular}{ll}
    \toprule[1.8pt]
    \multicolumn{2}{c}{\textbf{Results on Yelp}} \\
    \midrule
    Source & \textbf{decent} selection of meats and cheeses . \\
    Reference & the meats and cheeses were \textbf{not a lot} to choose from . \\
    Transferred & \textbf{limited} selection of meats and cheeses. \\
    \midrule
    \midrule
    Source & anyway , we got our coffee and \textbf{will not return} to this location .\\
    Reference & we got coffee and \textbf{we 'll think about going back} . \\
    Transferred & anyway , we got our coffee and \textbf{will definitely return} to this location. \\
    \midrule
    \midrule
    Source & everything we 've ever ordered here has been \textbf{great tasting} . \\
    Reference & everything we've ever ordered here has been \textbf{horrible tasting} . \\
    Transferred & everything we've ever ordered here has been \textbf{horrible tasting} . \\
    \midrule
    \midrule
    Source & \textbf{great} food, low prices , and \textbf{an authentic} mexican cantina vibe . \\
    Reference & \textbf{terrible} food , bad prices , \textbf{would not recommend} . \\
    Transferred & \textbf{terrible food}, low prices, \textbf{not an authentic mexican cantina vibe} . \\
    \midrule
    \midrule
    Source & i 'm \textbf{not willing} to take the chance. \\
    Reference & im \textbf{willing to} take the chance!  \\
    Transferred & i'm \textbf{going to} take the chance. \\
    \midrule
    \midrule
    Source & the evening started out \textbf{slow}. \\
    Reference & the evening started out with \textbf{excitement}. \\
    Transferred & the night came out \textbf{perfect}. \\
    \midrule
    \midrule
    Source & overall it was a \textbf{miserable evening} .\\
    Reference & overall it was an \textbf{exceptional evening} . \\
    Transferred & and it was a \textbf{great experience} . \\
    \midrule
    \midrule
    Source & the color that she uses on my girlfriend 's hair looks \textbf{great}. \\
    Reference & the color used on my friend was a \textbf{bad} choice. \\
    Transferred & the color that she put on my wife's hair looked \textbf{terrible}. \\
    \midrule
    \midrule
    Source & pricy but the cheese pies are \textbf{delicious} ! \\
    Reference & \textbf{pricy} and these cheese pies are \textbf{disgusting} ! \\
    Transferred & \textbf{pricy} and the cheese pies are \textbf{awful}! \\
    \midrule
    \midrule
    Source & it 's \textbf{not my fave} , \textbf{but it 's not awful} . \\
    Reference & it 's a very \textbf{pleasant surprise} . \\
    Transferred & it's \textbf{worth my fave}, \textbf{but it's definitely expensive} . \\
    \midrule
    \midrule
    Source & so far , \textbf{great customer service} . \\
    Reference & so far the customer \textbf{service was just rude} . \\
    Transferred & so disappointed, \textbf{very bad service} . \\
    \midrule
    \midrule
    Source & this place is a shit hole with \textbf{bad} service. \\
    Reference & this place is very \textbf{nice} with \textbf{great} service. \\
    Transferred & this place is a real \textbf{gem} with \textbf{great} food. \\
    \midrule
    \midrule
    Source & this branch is getting \textbf{worse and worse}. \\
    Reference & this branch is getting \textbf{better and better}.  \\
    Transferred & this place is getting \textbf{better and better}. \\
    \bottomrule[1.8pt]
    \end{tabular}%
      \caption{Examples of transferred results from RLM on the Yelp dataset. For every example, the first line is the input source sentence, the second line is the reference transferred output and the last line is the sentence generated by RLM. Stylization words are marked in bold.}
  \label{tab:YelpResults}%
\end{table}%

\begin{table}[htbp]
  \centering
    \begin{tabular}{ll}
    \toprule[1.8pt]
    \multicolumn{2}{c}{\textbf{Results on Amazon}} \\
    \midrule
    Source & then \textbf{both of them go into} the dishwasher . \\
    Reference & then \textbf{both of them might go into} the dishwasher .\\
    Transferred & but \textbf{none of them go into} the dishwasher.\\
    \midrule
    \midrule
    Source & i dropped phone once and the case \textbf{held up perfectly} .\\
    Reference & i dropped phone once and the case \textbf{didn't hold up}. \\
    Transferred & i dropped phone off and the case \textbf{slides off quickly} . \\
    \midrule
    \midrule
    Source & i \textbf{have many oxo products} ,  and i 've always been \textbf{pleased} . \\
    Reference & i \textbf{have a few oxo products} and have always been \textbf{disappointed} . \\
    Transferred & i \textbf{have used oxo products} , and i 've always been \textbf{disappointed} . \\
    \midrule
    \midrule
    Source & i use it a lot ,  and it \textbf{never gave me any problem} . \\
    Reference & i never use it a lot, it \textbf{always gives me problems}. \\
    Transferred & i tried it a lot, and it \textbf{never gave me a try}. \\
    \midrule
    \midrule
    Source & just do your homework and you will end up pretty \textbf{satisfied} . \\
    Reference & just do your homework and you will end up pretty \textbf{dismayed} . \\
    Transferred & just play your game and you will end up very \textbf{disappointed}. \\
    \midrule
    \midrule
    Source & it \textbf{stays on} and i unplug it when it s done . \\
    Reference & it won't \textbf{stay on} so i unplug it when it s done . \\
    Transferred & it \textbf{slides off} and i unplug it when it s used .\\
    \midrule
    \midrule
    Source & i have been missing out !  this thing is so \textbf{sharp} .\\
    Reference & i have not been missing out ! this thing is so \textbf{dull} .\\
    Transferred & i have been trying out! this thing is very \textbf{bad} .  \\
    \midrule
    \midrule
    Source & i have to say that it was \textbf{money well spent} . \\
    Reference & \textbf{money wasted} i think . \\
    Transferred & i have to say that it was \textbf{very soon died} . \\
    \midrule
    \midrule
    Source & \textbf{well worth} the money ,  \textbf{wouldn' t want} to be without it . \\
    Reference & \textbf{not worth} the money, \textbf{could easily} be without it. \\
    Transferred & \textbf{not worth} the money, \textbf{didn' t want} to play with it. \\
    \midrule
    \midrule
    Source & i \textbf{like it better than any} of the forman items . \\
    Reference & I \textbf{dislike it more then any} of the forman items . \\
    Transferred & i \textbf{like it better than none} of the forman items . \\
    \midrule
    \midrule
    Source & overall i \textbf{love} them ,  and would \textbf{probably buy them again} . \\
    Reference & overall i \textbf{hate} them , and would \textbf{never buy them again} . \\
    Transferred & unfortunately i \textbf{hate} them, and would \textbf{not buy them again}.\\
    \midrule
    \midrule
    Source & what \textbf{a change} in my life this will make . \\
    Reference & what \textbf{a bad change} in my life this will make . \\
    Transferred & what \textbf{a problem} in my life this will make . \\
    \midrule
    \midrule
    Source & i give it \textbf{five stars} for \textbf{making eating good so easy} ! \\
    Reference & i give it \textbf{four stars} for \textbf{making eating well more easy} ! \\
    Transferred & i give it \textbf{two stars} for \textbf{making eating good not easy} ! \\
    \bottomrule[1.8pt]
    \end{tabular}%
      \caption{Examples of transferred results from RLM on the Amazon dataset. For every example, the first line is the input source sentence, the second line is the reference transferred output and the last line is the sentence generated by RLM. Stylization words are marked in bold.}
  \label{tab:AmazonResults}%
\end{table}%

\section{Learning Objectives}

As described in Section~\ref{sec:implement-details-RLM}, we can use the variational mutual information estimators to bound the MI terms in the objectives:
\begin{align}
    I(\vx; \bar{\vs}, \vc_i) &\geq \mathbb{E}_{P(\vx_i, \bar{\vs}, \vc_i)} [\log P_\text{RLM}(\vx_i| \bar{\vs}, \vc_i) ] - \mathbb{E}_{P(\vx_i)}[\log P(\vx_i)] = \gL_1, \\
    I(\bar{\vs} ; \vc_i) &\leq \mathbb{E}_{P(\vs, \vc_i)}[\log Q(\bar{\vs}| \vc_i)] - \mathbb{E}_{P(\vs)P(\vc_i)} [\log Q(\bar{\vs}| \vc_i)] = \gL_2, \\
    I(\vx_i, \vc'_i) &\leq \mathbb{E}_{P(\vx_i, \vc'_i)} [\log P_\text{RLM}(\vx_i | \vc'_i)]  - \mathbb{E}_{P(\vx) p(\vc'_i)} [\log P_\text{RLM}(\vx_i | \vc'_i)] = \gL_3,
\end{align}
With training samples $\{(\vx_u, \bar{\vs}_n)\}_{u=1}^{U}$, we can estimate $\gL_1, \gL_2, \gL_3$ with Monte Carlo estimation:
\begin{align}
\hat{\gL}_1 &=  \frac{1}{U}\sum_{u=1}^U [\log P_\text{RLM}(\vx^u_i| \bar{\vs}^u, \vc^u_i) ], \label{eq:appendix-max-loglikeli}\\
\hat{\gL}_2 &=  \frac{1}{U}   \sum_{u=1}^U \log Q(\bar{\vs}^u| \vc^u_i) - \frac{1}{U^2}
\sum_{u=1}^U \sum_{v=1}^U \log Q(\bar{\vs}^v| \vc_i^u), \\
\hat{\gL}_3  &= \frac{1}{U} \sum_{u=1}^U\log P_\text{RLM}(\vx_i | \vc'_i)]  - \frac{1}{U^2} \sum_{u=1}^U \sum_{v=1}^U \log P_\text{RLM}(\vx^v_i | \vc'^u_i)].
\end{align}
In \eqref{eq:appendix-max-loglikeli}, we remove the term $\mathbb{E}_{P(\vx_i)}[\log P(\vx_i)]$, which is a constant with respect to the model parameters. Note that $Q(\bar{\vs} | \vc)$ can be regarded as a prediction to style $\bar{\vs}$ based on the content embedding $\vc$. Therefore, we build $Q(\bar{\vs}| \vc)$ as a linear style classifier, trained by the likelihood maximization:
\begin{equation}
\max_{Q}   \hat{ \gL}_\text{Var} = \frac{1}{U}\sum_{u=1}^U Q(\bar{\vs}^u| \vc_i^u).
\end{equation}


\end{document}